# Mediation Challenges and Socio-Technical Gaps for Explainable Deep Learning Applications

**R.R.M. Brandão**[1]  RMELLO@BR.IBM.COM
**J.L. Carbonera**[2]  JOEL.CARBONERA@INF.UFRGS.BR
**C.S. de Souza**[3]  CLARISSE@INF.PUC-RIO.BR
**J.S.J. Ferreira**[1]  JJANSEN@BR.IBM.COM
**B.N. Gonçalves**[1]  BNG@BR.IBM.COM
**C.F. Leitão**[3]  CFARIA@INF.PUC-RIO.BR

[1] *IBM Research*
*Rio de Janeiro, RJ 22290-240*

[2] *Informatics Institute, Federal University of Rio Grande do Sul – UFRGS*
*Porto Alegre, RS 91509-900*

[3] *Informatics Department, Pontifical Catholic University of Rio de Janeiro – PUC-Rio*
*Rio de Janeiro, RJ 22451-000*

## Abstract

The presumed data owners' right to explanations brought about by the General Data Protection Regulation in Europe has shed light on the social challenges of explainable artificial intelligence (XAI). In this paper, we present a case study with Deep Learning (DL) experts from a research and development laboratory focused on the delivery of industrial-strength AI technologies. Our aim was to investigate the social meaning (i.e. meaning to others) that DL experts assign to what they do, given a richly contextualized and familiar domain of application. Using qualitative research techniques to collect and analyze empirical data, our study has shown that participating DL experts did not spontaneously engage into considerations about the social meaning of machine learning models that they build. Moreover, when explicitly stimulated to do so, these experts expressed expectations that, with real-world DL application, there will be available mediators to bridge the gap between technical meanings that drive DL work, and social meanings that AI technology users assign to it. We concluded that current research incentives and values guiding the participants' scientific interests and conduct are at odds with those required to face some of the scientific challenges involved in advancing XAI, and thus responding to the alleged data owners' right to explanations or similar societal demands emerging from current debates. As a concrete contribution to mitigate what seems to be a more general problem, we propose three preliminary *XAI Mediation Challenges* with the potential to bring together technical and social meanings of DL applications, as well as to foster much needed interdisciplinary collaboration among AI and the Social Sciences researchers.

## 1. Introduction

Explainable Artificial Intelligence (XAI) has been rapidly growing into a multi-disciplinary and multi-perspective research field. Recent European regulation on the rights and obligations of personal information owners, controllers and processors (EUGDPR, 2016) has stirred the debate about what the European citizens' *right to explanation* may mean. The initiative has been followed by countries outside Europe, for both social and economic reasons. Regardless of whether legislation can or will go as far as to require explanations about how large datasets built with billions of pieces of *personal* information are used by machine learning applications, the stage is set for stronger societal questioning of what AI systems, especially those based on Deep Learning (DL) techniques, do and why (Goodman & Flaxman, 2016; Larus et al., 2018). DL systems, following the tradition of neural network systems, are mostly viewed as inscrutable *black boxes*, driven by sophisticated mathematical models operating on data patterns that do not necessarily (and frequently do not at all) have any correspondence with humanly meaningful patterns of information. Yet, the *social meaning* of AI, that is, what AI means *to others*, is a topic that our research community is now pressed to investigate.

In this paper, we present the findings and conclusions from a case study focused on the social meanings of DL models, as perceived by a group of experts from a research and development laboratory focused on the delivery of industrial-strength AI technologies. Our motivation for this work sprang from the apparent gap separating DL researchers and developers from real-world contexts of use and user requirements. Miller and co-authors (2017), for example, say that DL experts are looking for (and looking at) explanations and model interpretations *for themselves*. Much of the research about how to interpret and explain AI behavior, they say, is driven by the needs of those who *build* AI, and not necessarily of those who *use* it. This view has been recently strengthened by an extensive study of 289 primary papers and 12,412 citing publications covering XAI, interpretable Machine Learning (iML), and related areas. In this study, Abdul and co-authors (2018) verified that "unlike with intelligent and ambient systems, where there is a strong emphasis on validation with real users and scenarios, validation for [fairness, accountability and transparency of algorithms] and iML appear to be primarily performed on commonly available datasets." (p. 8) By working with expert ML professionals from an R&D lab, we were able to look for further evidence of related problems and their consequences, in a group whose projects habitually involve *real-world users*.

Our starting point was one of the very basic requirements of *good explanations*: that the explainer understands what the explanation *means* to the person who asks for it. Using qualitative research techniques to collect and analyze empirical data (Blandford, 2013; Blandford et al., 2016; Creswell et al., 2007; Denzin & Lincoln, 2005; Seidman, 2013; Yin, 2003), our study has shown that participating DL experts did not spontaneously engage into considerations about the social meaning of machine learning models that they build, even when such models are to be used in richly contextualized and familiar DL applications. Furthermore, when explicitly stimulated to do so, these experts expressed expectations that, in real-world DL applications, there will be available mediators to bridge the gap between technical meanings that drive DL work, and social meanings that AI technology users assign to it. By relating what we heard from our study's participants with evidence collected in specialized DL media and publications, we concluded that current research incentives and values guiding the participants' scientific interests and conduct are at odds with those required to face some of the scientific challenges involved in advancing XAI, and thus responding to the alleged data owners' right to explanations or similar societal demands emerging from current debates. As a concrete contribution to help mitigate what seems to be a more general problem, we propose three *XAI Mediation Challenges* that, we believe, have the potential to bring together technical and social meanings of DL applications, as well as to foster the required interdisciplinary collaboration among researchers in AI and the Social Sciences.

The remainder of this paper is structured in five additional sections. The second one presents a commentary on related work. The third section introduces the case study, with details on participants and methodology. The fourth one presents the results of the study. The fifth section discusses the scientific contributions and limitations of this work. Finally, the last section of the paper summarizes the study, highlights and contextualizes its results and contributions, and comments on our next steps. We also include a series of appendices with detailed information that readers may wish to consult.

## 2. Related Work

Explainable AI is not a new topic. Intelligent systems' ability to explain and justify their behavior has been a major concern for researchers. One of the frequently invoked reasons to invest in XAI is the end users' trust in AI systems, which is intrinsically related to their ability to elaborate a *compatible* conceptual model of how AI works. The notion of compatibility is important because, although users do not have to know exactly how AI works, they must conceptualize it in such a way that misunderstandings that lead to interactive breakdowns – or, worse, to unsuspected misinterpretations of AI results – are avoided. There is a large body of work about explanations for symbolic AI, which predominated in the 1980's. Some examples are Schank's *explanation patterns* (1986), focusing on cognitive modeling, Moore's (1994) and Cawsey's (1992) work on explanatory dialogues during interaction with knowledge-based systems, as well as Hovy's work on text generation under pragmatic constraints (Hovy, 1987, 1990). With the *winter* of expert systems and the advent of the semantic web, much of this work was redirected to the broader field of *question-answering* (Maybury, 2004), which typically involved non-verbal modes of interaction and representation. The evolution of this research to account for explanations and question answering with big data and machine learning applications naturally followed, and now stands among the hot topics in both AI and HCI.

On the AI side, a recent cursory survey of explanation and justification in machine learning by Biran and Cotton (2017) emphasizes the long history of explanations and justifications in AI. The survey separates work on interpretation and justification of predictions, i.e. the system's output from model interpretability, which has to do with understanding the system's inner configurations and parameters. For end users, the former is more relevant than the latter. Interpretability, however, is not a clear-cut concept. For this reason, Lipton (2016), for example, proposes to define interpretability in terms of desiderata, model properties and methods. His main point is that the problem of interpretability (and we can infer that of explainability too, although he does not mention this term in the paper) is still insufficiently defined, which by necessity partially compromises the solutions proposed so far. Moreover, Lipton establishes a connection between technical and social perspectives on the problem, calling for more 'critical writing' about the topic in the ML community. In his words, "the responsibility to account for the impact of machine learning and to ensure its alignment with societal desiderata must ultimately be shared by practitioners and researchers in the field." (p. 8) A similar call for reflection is voiced by Miller (2017), who urges XAI researchers to get acquainted with the long tradition of explanation studies in the social sciences and incorporate this knowledge in their research.

As a transition between AI and HCI, the work with explanations for recommendation systems has made comparatively more progress than other ML applications. This is possibly due to what Grudin (2009) identifies as a platform convergence. According to the author, in the past, research in AI and HCI had different time scales and goals. AI research was typically long-term and more ambitious, while HCI research was short-term and geared towards promoting interactive innovation and improved user experience. When ML techniques began to be incorporated in numerous kinds of popular applications, intelligent systems' usability jumped to the forefront in HCI researchers' agenda. Zhang and Chen's (2018) survey of explanations in recommendation systems shows that there are already several strategies being used, like user-based, item-based and content-based explanations, which can be expressed textually, visually, or in combined form. Of course, recommender systems can use different sorts of ML techniques. When DL is used, however, the authors verified that "in most cases, the recommendation/explanation model is still a black box and we do not fully understand how an item is recommended out of the other alternatives. This is mostly due to the fact the hidden layers in most deep neural networks do not possess certain understandable meanings." (p. 63) The problem begins with the notion of *understandable meaning*. Understandable to whom? One of the examples of cited work in Zhang and Chen's survey – Koh and Liang's (2017) approach to open the black box of DL systems by looking at training data – confirms Miller and co-authors' (2017) judgment that much of the effort in XAI is being dedicated to help DL *experts* (and not end users) understand what runs inside the so-called *black box*. This is neither new, nor necessarily a problem with research in explanations. The use of explanations to help developers debug and test systems has been around for a long time (e.g. Brusilovsky, 1994; Kulesza et al., 2010; Kumar, 2002; Wilson et al., 2003; Zacharias, 2009). Therefore, efforts like Pei and co-authors' (2017) to *whiten the DL black box* with a framework to overcome the currently heavy dependency of DL systems on manually labeled data that often fails to expose erroneous behaviors for rare inputs is also relevant to increase the quality of DL applications, and hence benefits the users. The only problem with this approach would be to attend the needs of DL experts and developers in detriment of DL applications' users.

Following Grudin's (2009) view of opportunities for joint research in AI and HCI, Dix (2016) and Abdul et al. (2018) report on challenges that current intelligent systems or AI enabled applications present to end users. Deep neural network models pose even harder explainability challenges when compared to other machine learning techniques such as Bayesian networks used in association with causal models (Pearl & Mackenzie, 2018), or DL models themselves used in association with prototypes (Li et al., 2017). User-centric AI explanations, which constitute the hallmark of HCI-inspired approaches, necessarily take into account the users' goals, context, abilities, knowledge and technological infrastructure. These are typical usability dimensions for HCI in general, as well as for AI-enabled HCI. A large part of them are essentially *pragmatic* dimensions, which is perfectly aligned with deeper considerations about the pragmatic nature of explanations.

Indeed, the importance of the pragmatic aspects of explanations transcends the realm of computer technologies. For example, in his discussion of *scientific* explanations, Faye (2014) underlines the fact that "any evaluation of the relevance of an explanation depends on both the content of the *explanans* and pragmatic factors concerning what is tacitly or implicitly understood in presenting a particular explanation in a particular situation. […] simply being true will not suffice to make it an appropriate explanation." (p. 183) This leads us to the discussions triggered by recent European regulation for data protection (EUGDPR, 2016), involving several research communities, like HCI (Veale, Binns, & Kleek, 2018), AI (Eiband, Schneider, & Buschek, 2018; Goodman & Flaxman, 2016), Law (Edwards & Veale, 2017; Selbst & Powles, 2017; Wachter, Mittelstadt, & Floridi, 2017), and Philosophy (Binns, 2017). In general, these discussions turn around the *letter vs. the spirit of the law*, or soft law, approved by the European community, as well as the much broader concerns about the impact of increasingly pervasive ML-enabled technologies. One of the most comprehensive analyses of such concerns is presented in a technical report sponsored by Informatics Europe, the ACM Europe Council, and the ACM Europe Policy Committee (EUACM), with European recommendations on machine-learned automated decision making (Larus et al., 2018). The report presents a set of ten technical, ethical, legal, economic, societal, and educational recommendations. The holistic perspective adopted by the eleven report authors from several research institutions and universities across Europe shows that we are facing an extremely complex task, which requires intensive disciplinary and interdisciplinary efforts.

Our specific approach chooses to work with a group of DL researchers who are habitually engaged in the development of real-world application. This choice is not new. Just a few years ago, Hill and co-authors (2016) have conducted a study with a group of expert ML developers. Unlike ours, however, their focus was set on another critically important aspect of trustworthy ML applications, repeatable software engineering processes. The authors concluded for the need to advance the state of the art in software engineering to address specifically the *trials and tribulations* of ML applications developers.

In this paper, we now look at how a similar group of developers from an R&D lab relates to the values and meanings of potential users of a realistic DL-enabled system. To the best of our knowledge, this kind of research has no precedent in XAI efforts directed to contemporary DL applications. As the details of our research presented in the next section will show, we are exploring this group's *positional thinking*. Some years ago, Nussbaum (2010) advocated for the importance of developing a child's *empathy* throughout K-12 education. Her argument is centered on the importance of literature and other subject matters related to the humanities to educate citizens who are able to exercise mutual understanding tolerance, both at individual and societal levels. Nussbaum defines positional thinking as "the ability to see the world from another creature's viewpoint". Our interest in probing a group of DL experts' positional thinking is not only related to socially responsible technology development called for by many of the works cited above, but also by Schank's (1986) earlier ideas on explanations. He distinguished between three positions that one can occupy in a spectrum of *understanding*, an indisputable pre-requisite for explaining. At one end of the spectrum is *making sense*, which Schank describes as "the point where events that occur in the world can be interpreted by the understander in terms of a coherent (although probably incomplete) picture of how those events came to pass." (p. 6) At the other end is, interestingly, *complete empathy*, defined as "the kind of understanding that might obtain between twins, very close brothers, very old friends, and other such combinations of very similar people." (p. 6) The use of *complete* as a qualifier of *empathy* is noteworthy, since it allows us to think that along the spectrum from end to end there are *less than complete empathy* points, which nevertheless refer to an *empathetic* explainer's positional thinking, as opposed to a predominantly rational perspective (let alone a self-centered one). Moreover, an investigation touching on aspects of ML or DL experts' empathy with the users of contemporary AI technology has the potential to contribute not only to XAI, but also to the study of algorithms accountability and technology-encoded social discrimination (Diakopoulos, 2016; Eubanks, 2018; O'Neil, 2016).

## 3. Case Study Participants and Methodology

Because our aim was to make an in-depth exploration of *meanings*, we adopted qualitative research methodology in this project (Blandford, 2013; Creswell et al., 2007; Denzin & Lincoln, 2005). We chose to carry out a case study (Yin, 1981, 2003), a typical option when the aim of research is to answer "how" or "why" questions in exploratory investigations. Their distinguishing characteristic is to examine a contemporary (not historical) phenomenon in its actual (real-life) context of occurrence, a crucial feature when the boundaries between the phenomenon and its context are not clear, and the researcher cannot clearly identify and isolate variables. As a research strategy, case studies enable an intensified manifestation and rich analysis of situated meanings within the boundaries of one or more clearly identifiable units of analysis. The research question we wanted to answer was: How do Deep Learning researchers whose R&D lab work involves the development of real-world applications *relate* to the social meaning of their expert work?

Our research team is a cross-disciplinary one, including members with doctoral degrees in the Humanities, Social Sciences, and Computer Science. As a group, we have research experience not only in AI (three of us), but also in Human-Computer Interaction (four of us). We started our study with a direct invitation to all DL experts of a medium-sized R&D lab (with 60+ researchers, from several expert areas) to participate in the study. The requirement for participation was to have work experience with ML, Neural Networks (NN) or Deep Learning. Fifteen of the invited members (12 males, 3 females) agreed to participate. In Figure 1, we show information about their background and experience with ML in general (including NN and DL). The vast majority of the participants (80%) either held a doctoral degree (67%) or were in the course of their PhD studies (13%).

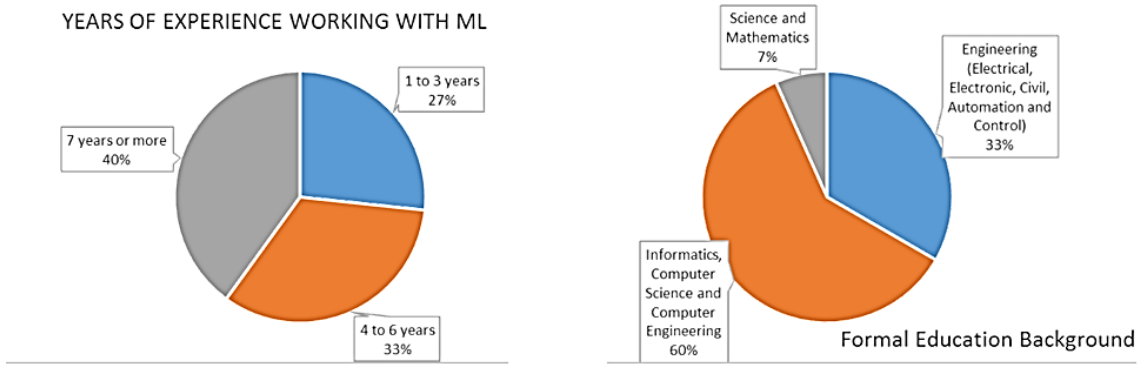

Figure 1: Participant's background and experience

Our research project started with a *survey* (see Appendix 1 – Survey Questions), including closed and open-ended questions. In addition to demographic information, we asked the participants about their **motivation** and **personal history** with ML, NN or DL. We also asked them about the **tools** they used in their projects, with special attention to **pros and cons** they saw in such tools, and the reasons why they used it. The data collected with this survey was analyzed quantitatively for frequency and qualitatively for recurring themes (Braun & Clarke, 2006; Creswell & Clark, 2017). This information allowed us to identify our *case* and design the *study*.

The case focused on with a subgroup of seven participants, with diverse backgrounds and professional experience but with intersecting technical skills. All of them used Keras (Chollet, 2017; Keras Documentation, 2018) to develop DL models. The initial step in the study was to run a pilot test to check data collection procedures. In order to elicit the social meanings that our participants assigned to DL models they built, we designed a carefully contextualized *hands-on* activity with Keras, followed by an in-depth *interview* with open-ended questions (Seidman, 2013).

The **context** for the hands-on activity and interview was very familiar to participants. It consisted of a hypothetical, though realistic, DL application to recognize electoral votes in paper ballot forms manually filled up by voters. For an existing application of the same sort, see the work of Ji et al. (2011). The application was to be used as a backup for the electronic voting system used in the participants' own country (which, as voting citizens, they knew very well). The details of this context can be found in Appendix 2 – Scenario and Hands-On Activity).

In our initial design, we planned to: (1) present the scenario to the participant; (2) ask the participant to build a basic (possibly incomplete) DL model for the application, using Keras and the MNIST dataset (or any other dataset of their choice); and (3) run the interview. The interview was structured in three sequential subsets of questions: (a) the participant's spontaneous comments and reactions to the proposed scenario and activity; (b) the participant's comments on significant parts or aspects of their model and the code they used to build it; and (c) the participant's comments on aspects of the development and use of the proposed technology (see Appendix 2 – Scenario and Hands-On Activity for details).

After two pilot tests with different participants, we adjusted the data collection procedures and the time of the interview. We thus carried out the actual study with five remaining participants: four PhDs

| Research Phases | Survey | Case Study | | |
|---|---|---|---|---|
| | | Pilot Test 1 | Pilot Test 2 | Final Hands-on Activity and Interview |
| Participants | 15 | 1 | 1 | 5 |
| | | 7 | | |
| Total Number of Participants | 15 | | | |

Table 1: Distribution of participants throughout the research project.
A total of fifteen machine learning experts participated in the study, five of them contributed to our final case study with hands-on activities followed by interviews.

– three of them with 7+ years of experience and one with 4-6 years of experience – and one PhD candidate with 1-3 years of experience with DL. In Table 1, we show the distribution of participants throughout the entire research project.

The focus of our case study analysis was the content of the five interviews, which lasted on average 47 minutes each and were all video-recorded and transcribed. We used thematic analysis techniques (Braun & Clarke, 2006) to find out the social meanings that our interviewees assigned to DL models in a real-world application scenario. To ensure the rigor and validity of our research, we followed the long process depicted in Figure 2, using the qualitative analysis support software QDA Miner (Provalis, 2018). On top of Figure 2 (region 1) are the *tuning* steps to consolidate the set of *codes* and *categories* to be used in the entire process. Three of us independently coded the longest interview. We then compared our *codebooks* and coding strategies and produced a revised set of coding *categories* (or themes). One of the three independent coders (coder 1) was then in charge of reanalyzing the same interview, using the revised *codebook*. The result of this analysis was discussed and validated with the other two independent coders. Thus, the *tuning* steps were completed, and the thematic analysis proceeded for all interviews. At this second stage of the process (region 2), coder 1 iteratively analyzed the five interviews, reaching for higher-levels of abstraction and articulation of themes and thematic relations. Results of the analysis were then internally validated with peer debriefing (Yin, 2003), which allowed us to consolidate the conclusion of the case study analysis. The final stage of the process (region 3) consisted of an external validation through member checking (Yin, 2003), i.e. the verification of conclusions with the participants of the study themselves. In a seminar with all five interviewees, we presented the methodology, the results and the conclusions of our study, in anonymized form.

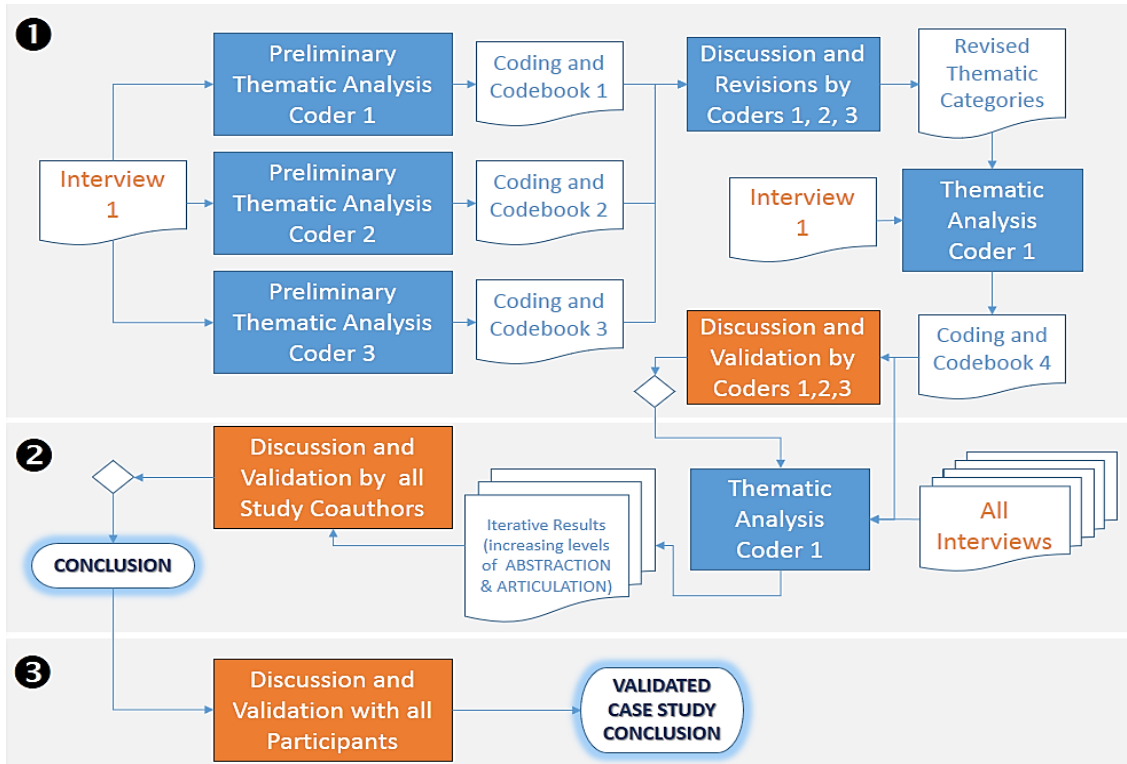

Figure 2: Procedural steps in the analysis and validation of the case study interviews

## 4. Results

As mentioned in the previous section, the results of the *survey* were used mainly to help us make decisions about the sampling and scenario of the case study. We therefore start with a more concise presentation of *survey* results, and then proceed to a detailed presentation of *interview* results.

### 4.1 Survey Findings

The demographic information from the survey has been presented in the previous section (see Figure 1). Answers to the closed-ended question about **tools** used by the 15 participants showed that 66.7% of them used TensorFlow, 53.3% used Keras, 40% used Caffe or other tools and 20% of them used neither TensorFlow, nor Keras. Therefore, our first finding was that TensorFlow was the most used model-building tool in our sample. The second most used tool was Keras.

Answers to the open-ended questions were thematically analyzed by one of us and validated with peer briefing. In close correspondence with the topic of the questions, the top-level themes in the *survey* were: **motivation**, **domain (of specialization)**, **training**, and **technology**. Each one of these themes included content related to sub-themes. Because the questions and answers were very straightforward, we do not include examples of the participants' answers. Instead, we show in Figure 3 the themes, sub-themes and their frequency of occurrence (i.e. the percentage of 15 participants whose answers included content related to that theme). *Case* (see top of the image in Figure 3) is the term used by our qualitative analysis software to refer to a unitary source of evidence. In our *survey*, this unitary source is a single form filled by a single participant. Hence, in addition to a relative count (*% Cases*), we also see the absolute count (*Cases*) of participants who mentioned content related to the listed themes and sub-themes.

Regarding **motivation**, we found that nearly all participants (with one exception, only) mentioned that their motivation was associated with the fact that DL helps them solve problems (may explicitly mentioned *real-world* problems) and find applications for machine learning. Additional motivations mentioned by some of the participants included professional development and curiosity about how to encode ML models. This finding was not surprising for an industrial R&D lab, whose hiring strategies naturally select researchers who are explicitly interested in applying research results to real-world applications with potential product developments.

MEDIATION CHALLENGES AND SOCIO-TECHNICAL GAPS FOR EXPLAINABLE DEEP LEARNING APPLICATIONS| | Cases | % Cases |
|---|---|---|
| **Motivation** | | |
| • Additional Motivation | 2 | 13.3% |
| • Mathematical Modeling, Science | 6 | 40.0% |
| • Problem Solving, Applications | 14 | 93.3% |
| **Training** | | |
| • University Degree Program | 11 | 73.3% |
| **Other than University Degree** | | |
| • In Job Training / Peers | 4 | 26.7% |
| • Online Training | 4 | 26.7% |
| • Self Taught / Self Initiative / Documentation | 12 | 80.0% |
| **Technology** | | |
| **Software Development** | | |
| • Ease of Use | 11 | 73.3% |
| • Programming Capabilities | 9 | 60.0% |
| • Range and Flexibility | 7 | 46.7% |
| • Model / Program Libraries | 5 | 33.3% |
| • Cost/Benefit | 3 | 20.0% |
| • Open Source or Proprietary | 3 | 20.0% |
| **Support and Documentation** | | |
| • Software Documentation | 6 | 40.0% |
| • Engaged Online Community | 6 | 40.0% |
| • Recommendation | 1 | 6.7% |
| **"Work in Progress"** | | |
| • Constantly Evolving | 3 | 20.0% |
| • Good for Research | 5 | 33.3% |
| • Performance | 2 | 13.3% |
| **Domain** | | |
| • Image Processing | 5 | 33.3% |
| • Pattern Recognition | 3 | 20.0% |
| • Remote Sensing | 2 | 13.3% |
| • High Performance Computing (HPC) | 1 | 6.7% |

Figure 3: Screenshot of QDA Miner, with themes, sub-themes, count and frequencies (where case = participant)

Regarding **domain (of specialization)**, our participants' – also not surprising – specialize in those that are relevant for their lab's activities. Because we did not include an explicit question about their domain of specialization, not all of them mentioned it in any of their answers. Thus, the manifestation of this kind of content was spontaneous, which underlines the significance of the theme whenever it occurred. Note that the level of abstraction of themes is not the same. While the most frequent domain theme was image processing, we have also detected the presence of pattern recognition (which is an important aspect of image processing and remote sensing, for example).

Regarding **training**, our findings revealed that, by far, the most frequent form to learn (get training) in DL is the participant's own initiative, looking for scientific and technical publications, software documentation, online forums and examples, blogs, etc. The second most frequently mentioned source of training was formal education in colleges and universities. Note that, as mentioned in the previous section, 80% of the participants either had a PhD degreed or were enrolled in a PhD program.

Because we had several explicit questions about technology in our *survey* (see Appendix 1 – Survey Questions), **technology** was the richest of all themes in terms of the variety of subthemes it encompassed. When talking about the pros and cons of the technologies they used, as well as about the reasons why they chose to use them, participants touched on software development, support and documentation, and the fact that DL technologies are – as of today – "a work in progress". The most recurring subthemes were all related to software development (ease of use, programming capabilities, range and flexibility).

Our participants' **motivation** confirmed that this group was appropriate for an investigation of *social meanings* of DL technology. They were remarkably motivated to deal with real-world applications. For example, P8 said that his motivation was DL's "*application to real world problems, specifically modeling and controls*". Likewise, P9 mentioned that her motivation came from "*creating powerful tools for building predictive models given recorded data, which are useful in many practical applications.*" We thus decided that the case study should involve a *hands-on* (practical) activity followed by an *interview*. The *hands-on* activity should support a technologically and socially contextualized topic for the *interview* with participants.

The participants' declared strategy to seek for **training** independently, looking for publications and online resources, as well as the importance of formal education programs, led us to choose one of the most popular examples used as an introduction to DL – the recognition of handwritten digits with the MNIST dataset. This choice was followed by the option for Keras as a technology. Although TensorFlow was the most used technology (66.7%) in the group, Keras was also used by more than half of the *survey* participants (53.7%). Moreover, they strongly valued ease of use (73.3%), which was recognized as one of the advantages of Keras over TensorFlow. P2 provides us with a particularly rich evidence that choosing Keras as a technology was a good option. He said that "*Keras and TensorFlow are more or less the same thing now. What happens it that Keras is an abstraction layer for Deep Learning using two different backends: TensorFlow and Theano (…). So basically, Keras is a high-level framework for a more complicated backend. In my opinion, TensorFlow is programmer-oriented, while Keras is ML-practitioner-oriented.*"

Given the decisions above, our last step with the *survey* was to elaborate a socially contextualized scenario, where Keras and MNIST could be used as a starting point for the in-depth *interview* to collect the data we needed. Our choice was to use the case study participants' own contemporary political context, with approaching general elections at the time of the interview. All of them are voting citizens of a country where a stable electronic voting system has been used for more than two decades. They are therefore thoroughly familiar with the system. We created a hypothetical realistic scenario for the study, in which participants would be part of the development team hired to develop a new **backup** system for the existing electronic voting system (EVS). According to their country's legislation, in case of EVS failure, the ballots must be manually written in printed forms. The processing of the ballots in such cases is also manual. The new backup system would be used to process the ballots *automatically*, using handwriting recognition techniques. Keras and MNIST came into this picture because voters must use their candidates' *numbers* in the ballots, not their names (see Appendix 2 – Scenario and Hands-On Activity for details). This choice was more than just a plausible ML application. For example, Toledo and co-authors (2015) discuss ML techniques to process voters' selections in the Australian ballot. Another example is Ji and co-authors' (2011) work in the context of Florida's Leon County ballots. In the latter case, the authors present several challenges for optical scan ballots, including "non-serious" write-in votes for candidates like Mickey Mouse, Harry Potter, Ronald McDonald, and others. This indicates that our scenario was perfectly in line with the *survey's* participants' R&D profile, with a pronounced interest in working with research challenges that could be used to solve real-world problems.

The results of the *survey* provided a solid base for the design our *case study*. Not only could we identify the scenario for the hands-on activity and interview, but also the subgroup of participants with shared professional values, technical knowledge and research interests.

### 4.2 Interview Findings

The findings reported in this section have been extracted from two of the three interview parts. In Appendix 3 – Interview Guide, we detail the interview guide and show the content for each part. Part 1 captured the respondents' spontaneous reactions to the proposed *hands-on* activity and scenario. Part 2 captured their comments on the Keras script they used to create a preliminary handwritten digit recognition model for the proposed application. All participants used MNIST; none of them asked about other datasets for the task (which they could have done, if they wished). Finally, Part 3 captured their views on the development and use of the proposed technology. Because the content collected in Part 2 was by comparison too technical, with little contribution to answer our research question, we worked only with the content from parts 1 and 3.

As mentioned in the section on participants and methodology, the data collected in the *interview* was analyzed using thematic analysis techniques (Braun & Clarke, 2006). The process was supported by qualitative data analysis software (Provalis, 2018), which also allowed us to track the frequency of themes in the data. The use of frequency must be handled with care in qualitative research. Vaismoradi and co-authors (2013), for example, propose that the main difference between content analysis and thematic analysis, which are often used as interchangeable labels for the same kind of analytical work, "lies in the possibility of quantification of data in content analysis by measuring the frequency of different categories and themes, which cautiously may stand as a proxy for significance." (p. 404) Although our analysis makes use of theme frequency, it will be shown that it is clearly a thematic one, given the crucial importance of explicit and implicit *meaning* relations that we detect among themes, at different levels of abstraction.

To facilitate the presentation, we will use the words *code* and *category* when referring to the lower levels of abstraction in the analysis, which are directly supported by QDA Miner. The word *theme* will be used to refer to higher levels of abstraction, when meaning categories are articulated with one another to produce the final answer to our research question. Therefore, the *codebook* used in the analysis (see Appendix 4 – Codebook) reveals the kinds of lower-level meanings we found in the transcriptions of parts 1 and 3 of the whole set of interviews. There are six categories in it:

- **Attitude and Relation:** This category was used for content related to the participant's attitude regarding the topic of the question/answer, or his relation with it.

- **Trust:** This category was used for content related to the participant's own, or others', trust and confidence in the topic of the question/answer.

- **Machine Learning:** This category was used for content related to technical, theoretical, or other aspects of ML.

- **Software Development and Stakeholders:** This category was used for content related to several aspects of the development and use of the system proposed in the scenario for the hands-on activity and interview.

- **Reaction to the Interview:** This category was used for content related to the participant's reaction to the interview itself, i.e. to being asked to answer the kinds of questions we asked, in the order and the way we asked them.

- **Perspective:** This category was used for content related to the participant's framing of the topic under discussion.

Each category encompasses a number of codes, the first-level of content classification in the analysis.

A comparison of code frequency is helpful to introduce interview findings and call the attention to important aspects in the interpretation of data and results. In Figure 4, we chart the frequency of all codes (organized by category) in Parts 1 and 3 of the interview. Regardless of numeric details, which can be found in Appendix 5 – Frequency of Codes and Categories in Part 1 and 3, total code frequencies in the second and third tier (dashed bounding box) show a significant phenomenon: in all but two categories – **Trust** and **Perspective** – there are pairs of inverted predominant frequency in the first and last part of the interview. More than the frequent meanings *induced* by the questions in Part 3, the significance of inverted frequency pairs lies in frequent meanings that the participants have (or haven't) introduced *spontaneously* in Part 1.

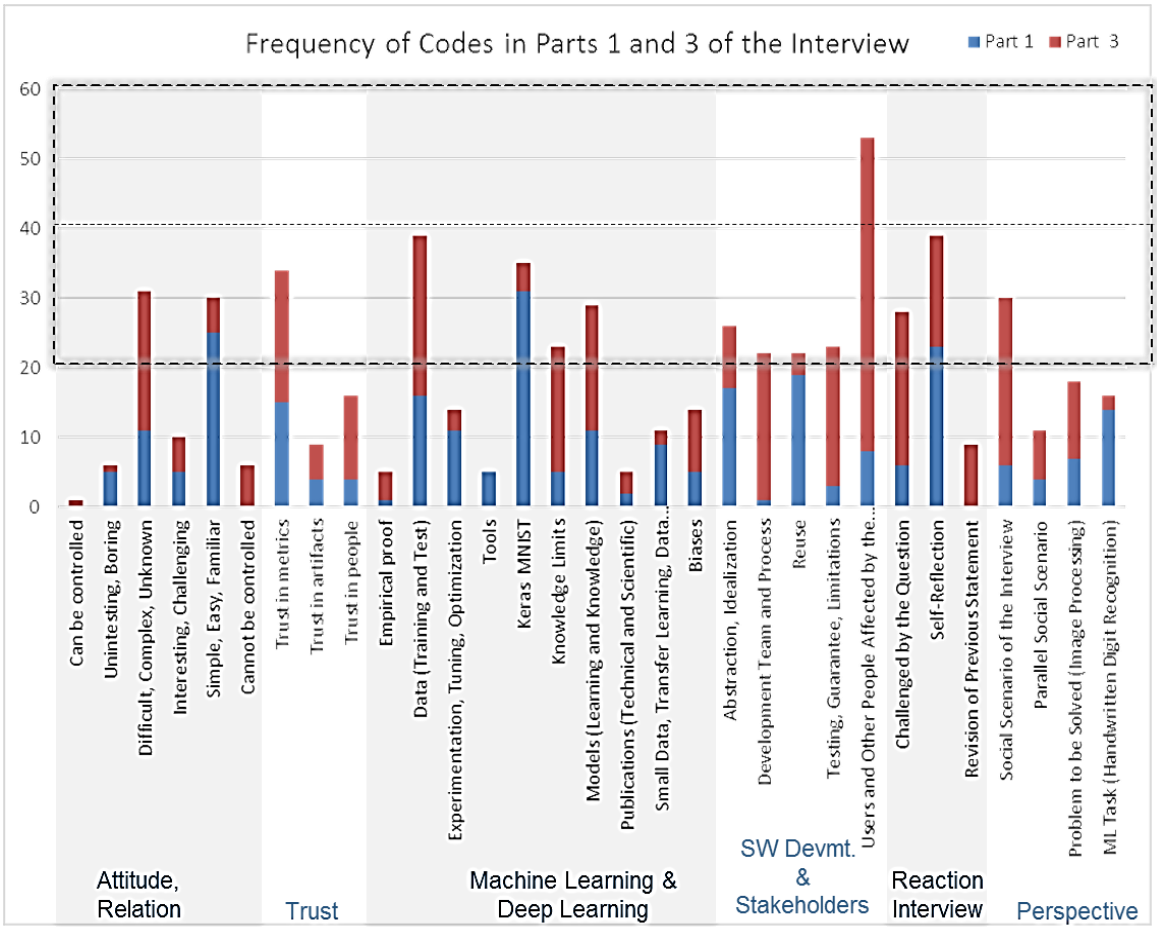

Figure 4: An overview of Code Frequency Contrasts in Parts 1 and 3 of the Interview

Starting with **Attitude and Relation**, whereas *Simple, Easy, Familiar* was remarkably more frequent in Part 1 than in Part 3, *Difficult, Complex, Unknown* was remarkably more frequent in Part 3 than in Part 1. These are *opposite codes*, of course, but this contrast is even more meaningful than would be the case if only one polarity had been coded. For example, if we had used only *Difficult, Complex, Unknown* as a code, we would be able to see that participants explicitly produced more content in this category in Part 3 than in Part 1. However, by coding both polarities, we have very strong evidence that the topics discussed in Part 1 were easier, simpler, more familiar to the interviewees than the topics discussed in Part 3 (which were more difficult, more complex, and less known). When discussing qualitative aspects in this section, we will see which topics these are. For now, it suffices to comment that in Part 1 of the interview participants were talking freely about their spontaneous reaction to the proposed *hands-on activity* of the case study, whereas in Part 3 they were necessarily talking about the proposed DL application, the recognition of handwritten ballot forms (see the Interview Guide in Appendix 3 – Interview Guide).

The next significant contrast in Figure 4 is between the prevalence of the *Keras, MNIST* code in Part 1, compared with the prevalence of *Knowledge Limits* in Part 3. Both codes belong to the **Machine Learning** category. The two other high frequency codes in this category (*Data* and *Models*) are also more frequent in Part 3 than Part 1, but the contrast is less striking than in the previous case.

The contrasts in the **Software Development and Stakeholders** category require further comments. Firstly, this is the only category where all the codes are higher-frequency codes (all codes, in all categories, included). Secondly, this is also the only category with one code in the very high frequency range (*Users and Other People Affected by the System*). The massive prevalence of three of five such codes in Part 3 of the interview is due to the topic under discussion: the third part of the interview was designed to explore the interviewees thoughts about the development and use of the DL application proposed in the *hands-on activity* scenario. Therefore, the significant contrast in this case is between the explicable massive frequency of *Development Team and Process*, *Testing*, *Guarantee*, *Limitations* and *Users and other People Affected by the System* in Part 3 with the prevalence of *Abstraction, Idealization* and *Reuse* in Part 1 (qualitative aspects of this contrast are discussed below).

In the **Reaction to the Interview** category we also have a sharp case of inverted predominant frequency with *Challenged by the Question*, most frequent in Part 3, and *Self-Reflection*, most frequent in Part 1. Given the topic of parts 1 and 3, it is not surprising that the participants were not challenged by questions in Part 1 (whose topic was their spontaneous reactions to the proposed material). It is however very significant that they *were* challenged by questions in Part 3. Also, it is not surprising that they were explicitly talking *less* about themselves (*Self-Reflection*) in Part 3, whose questions explicitly referred to *others*. These frequencies also underline the consistent correspondence between the design of our data collection instrument (see the interview guide in Appendix 3 – Interview Guide) and the analysis of themes brought about by the participants' response to the interview.

As a final remark on inverted predominant frequency pairs, there is only one mid-range frequency code in the **Perspective** category; the *Social Scenario of the Interview* is predominant in Part 3 (not surprisingly). However, this can be clearly contrasted with the lower frequency pattern of *ML Task* in Part 1, which is significant. It means that the participants' most frequent topic of conversation when talking about their spontaneous reaction to the material they had received prior to the interview was technical, rather than social.

The quantitative aspects of coding, as mentioned above, are not the focus of our analysis. Nevertheless, they already show some of the significant findings in this case study. Moreover, they help us clarify an important interpretive choice in the process of analysis, i.e. to focus on codes with mid and high-range frequency, using low-range frequency codes only to illuminate aspects of the former. The rationale for this choice is that we sought to capture the *collective discourse* of the case study's participants, that is, content that most or all of them contributed to our research. As a note on validation (see the next section), the confirmation of what we concluded to be this group's collective discourse naturally called for *member checking* (Yin, 2003) as validation, which we did.

### 4.3 Themes and Conceptual Maps

In this subsection, we articulate the relations among codes and elaborate the main *themes* of this case study. We start with the entire presentation of findings for Part 1, then Part 3. The connection of both parts is presented in a separate final subsection (4.4 Conclusions of the Study).

#### 4.3.1 Part 1

The frequent codes examined in Part 1 are listed below (most frequent first). We have selected them heuristically, looking for a break point in a frequency plot near half the value of the highest frequency (see details in Appendix 5 – Frequency of Codes and Categories in Part 1 and 3).

1. Keras, MNIST [Category: Machine Learning]
2. Simple, Easy, Familiar [Category: Attitude and Relation]
3. Self-Reflection [Category: Reaction to the Interview]
4. Reuse [Category: Software Development & Stakeholders]
5. Abstraction, Idealization [Category: Software Development & Stakeholders]
6. Data (Training and Test) [Category: Machine Learning]
7. Trust in Metrics [Category: Trust]
8. Task (Handwritten Digit Recognition) [Category: Perspective]

When talking about their spontaneous reaction to the material they received prior to the interview, the participants took a remarkably technical approach *in general*. We found evidence of all the codes above in the content produced by almost all the participants. The exceptions in this part of the interview are: P3, who did not give us evidence for *Self-Reflection* and *Data (Test and Training)*; and P5, who did not give us evidence for *Abstraction, Idealization*. Note that the content illustrated below has been produced in the participants' native language, which we translated into English for this publication.

We can illustrate content coded with *Keras, MNIST* in Part 1 with P1, who mentioned that "*This MNIST (data) base has been considered as solved for a long time now.*" P4 remarked that he was "*very familiar with Keras examples*". P3 said that "*for this problem, if you are only […] writing digits, MNIST is surely an interesting dataset. Especially for those who've seen Yan LeCun's presentations*".

*Simple, Easy, Familiar* content can be illustrated with P2, who said: "*Because I already know MNIST, [when I read the material] I sort of mapped directly what the possible solution would be.*" P4 mentioned that "*In the case of this particular problem, it's actually a very simple problem [to solve].*" P5 also said: "*I had the model in my head. I had virtually no work to do [for the proposed activity].*

*Self-Reflection* content touched on the participants' personal experience, beliefs and preferences, as in the case of P4, who mentioned: "*I have been working with small data problems. So, my mind is kind of set on this because of my research.*" P1 said: "*My method is always to begin with the simplest options.*" P2 remarked: "*when dealing with something that I don't know well enough, if I read [what has been published] about it, I will perhaps have a clue of how to improve my results.*"

For *Reuse*, the evidence we collected came from content like P1's: "*I already knew that there was a [convolutional] net working reasonably well for MNIST, so I did not [bother to] take the one with best results.*" P4 also said: "*[If] you can successfully draw a parallel between the problem you have and some other problem for which there are lots of data, then you can, indeed, think of approaches that will work reasonably well with little data, as long as you have available models trained with lots of data.*" P5 mentioned: "*With the Internet, these days, we are lazy with learning commands, and this sort of stuff. […] So, the first thing I did was to go to [the Internet] and search for a convolutional net using MNIST.*"

Content coded with *Abstraction and Idealization* was typically associated with coded content for *Reuse* and *Task*. In the paragraph above, we already see a clear example, with P4's statement referring to reusing ML models as an alternative to solving 'small data' problems. The problem context, in this case, is clearly abstracted or idealized. Another illustration of this comes from P2, who said: "*[For] this problem, this is known as the state of the art, the solution using DL. You cannot beat its accuracy with other techniques or other models. So, I would use it for this reason.*" Again, the specific context of DL application was abstracted away by the participant. The connection with content coded with a 'Task' perspective is also implicit in these examples. For instance, "the problem" as seen by P4 and P2 was not the situated ballot recognition one, in case of primary EVS failure (the *Social Scenario of the Interview*), but achieving a digit recognition task. An additional strong piece of evidence of *Task* perspective was produced by P3, who said: "*I read the scenario, but I was looking at keywords ... There! MNIST, Keras. This is what I kept in mind from the scenario.*"

Regarding content coded with *Data (Test and Training)*, we already have evidence in the examples above of how the participants talked about data (e.g. P4's views on small data and P3's comments on LeCun's achievements with the MNIST dataset). There are, however, additional aspects of the content referring to *Data*. For instance, P2 said: "*In general, I would start with the data that I have ... Since this is a case of supervised training, you must have the data to train your model. So, I think this is the first step to take: to access the data, and then look at the data.*" We should also emphasize P1's remarks about data, which unlike all other participants, raised a *social meaning* problem directly related to the scenario of the interview in Part 1. He said: "*But this issue in the scenario has an important aspect to it, the fact that handwritten digits are very regional, I mean. So, I guess that to be used with an EVS, MNIST would not be the best option. […] We'd have to create a dataset, I think… There are some available, like a dataset for handwritten checks, and all… But… Checks have these large numerical valued, and it's not just digits, right?*"

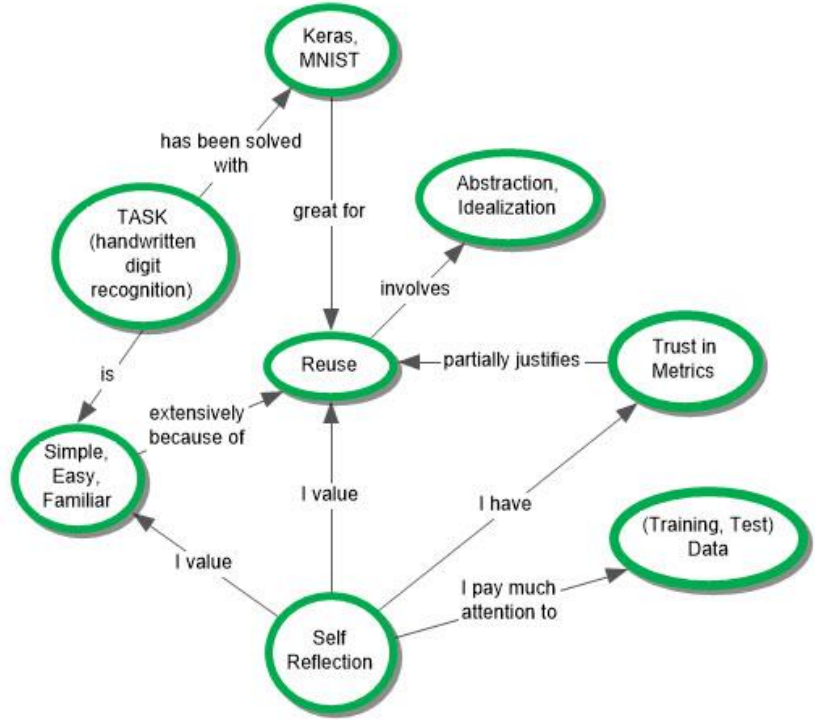

Figure 5: Higher order articulation of themes in Part 1

Finally, we had abundant evidence of content related to *Trust in Metrics* in Part 1. All participants invoked accuracy numbers to justify their choice of model, or to comment on the quality of a DL solution. For quick examples, we can start with P3's statement: "*Because we now have nearly 100% of correct answers for classification problems [like this], MNIST is always a good option, at least to begin with. Even like for drafting an architecture for character recognition.*" Additionally, we have P4 saying: "*I don't have to think about complicated models, since there is, already, a complete working solution with more than 99% of accuracy. It is a consolidated model. So, I don't have to worry.*"

The next step in our thematic analysis of Part 1 was to derive from the data consistent relations among codes (themes), looking for higher-order meanings. We refer to this process as the *articulation* of themes, whose result is presented as a conceptual map in Figure 5.

The justification for the labels used in the edges between nodes in Figure 5 is mostly present in the participants' discourse itself, which has been exemplified with excerpts in the paragraphs above. One of them, however ['Reuse' –involves→ Abstraction, Idealization] may require further justification. This is partly an implied meaning in content like P4's commentary on reusing models trained with *other* datasets ("*you can, indeed, think of approaches that will work reasonably well with little data, as long as you have available models trained with lots of [related] data*") or P3's comments on reusing LeCun's work with MNIST to draft convolutional net architectures for different image recognition tasks ("*MNIST is always a good option, at least to begin with. Even like for drafting an architecture for character recognition.*"). A stronger justification, however, comes from the technical characteristics of reusable software in general. Reuse is *intrinsically* related to abstractions, and abstractions necessarily involve idealization, that is, the decision to disregard specific (proper) features and distinctions real-world objects when creating computational representations for them.

MEDIATION CHALLENGES AND SOCIO-TECHNICAL GAPS FOR EXPLAINABLE DEEP LEARNING APPLICATIONS### 4.3.2 Part 3

The frequent codes examined in Part 3 are listed below (most frequent first). Just like with Part 1, we have selected them heuristically, looking for a break point in a frequency plot near half the value of the highest frequency (see details in Appendix 5 – Frequency of Codes and Categories in Part 1 and 3).

1. Users (or other affected parties) [Category: Software Development]
2. Social Scenario for the Interview [Category: Perspective]
3. Data (Training and Test) [Category: Machine Learning]
4. Challenged by the Question [Category: Reaction to question]
5. Difficult, Complex, Unknown [Category: Attitude, Relation]
6. Development Team and Process [Category: Software Development]
7. Tests, Guarantee, Limitations [Category: Software Development]
8. Trust in Metrics [Category: Trust]
9. Models (Learning and Knowledge) [Category: Machine Learning]
10. Knowledge Limits [Category: Machine Learning]

Evidence for *Users (or Other Affected Parties)* and *Social Scenario for the Interview* were abundantly provided in Part 3 of the interview, whose questions listed in Appendix 3 – Interview Guide explicitly addressed content of this sort. For example, one of the questions asked about *what BackSys, when implemented, could be expected to do well and not so well, and why*. Another question asked about *what kinds of doubts or issues BackSys users and other affected parties might have regarding the system's behavior, and why*.

For example, P2 said: "*The problem is that you would have statistics that are apparently OK for the candidates, I mean, that would make sense, but in fact the result you have would not reflect reality. This is a harder error to trap.*" Likewise, P4 mentioned the following: "*I would try to focus on [metrics], because [this] is the part that they are most interested in. They have no interest, I suppose, in understanding how the model works. But what they are greatly interested in is on whether they can trust the model, on how much they can trust in that model.*"

The other codes give us deeper and complementary insights into the participants' views on the social meaning of the DL component of BackSys, which they would be responsible for developing according to the proposed scenario. Starting with *Data (Training and Test)*, P1 said: "*I'd have to maximize the representativeness of the data. If I am going to use the system across the entire country, [to have data from] people in different social conditions, different levels of schooling, different regions. I don't know if people [here] have the same handwriting as in [other places] [...] It doesn't have to do with the model. It's extremely important, when you acquire your data, to have data that represent the real world.*" Compared to other participants, P1 was relatively more aware of *social meanings* in DL models and data, even in Part 1 of the interview (see comment on P1 on p. 15). Other participants had a different perspective on data. P4, for example, talked about the representativeness of training data in broader terms. He said: "*I'd need a training dataset, a set of ballots that could be used to know the patterns of possible errors [that BackSys should learn].*" P2 talked about dataset biases in more general terms, as well: "*The data you have used as input for the model is labeled by human beings, you know?*

*So, there is already an implicit bias there, the human error when labeling it. You would hope that the statistical law of large numbers can attenuate that.*" P3 also talked about biases but looking at how datasets are reused when task-specific datasets are not available: *"[...] the first layers of the convolutional net you built with ImageNet remain the same [...] the final layers of the net are changed with the dataset we have in hands. People are aware of biases, but they keep working with [ImageNet] anyway."*

Possibly because Part 3 of the interview induced a reflection on social meanings, all participants, who were spontaneously keen on technical aspects of DL, were at some point *Challenged by the Question* being asked. For instance, when asked about what kind of system behavior might be surprising or raise people's suspicion, P1 said: "*Hmmm… I can't tell… [Pause] It all depends on how it works… Unless, I don't know, suppose that some information leaks, because… [Pause] To me, it's a black box. You are going to feed in the ballots, the system will recognize [the votes]. And because there is a large volume [of ballots], it's kind of hard to see what the system is doing.*" P2 said: "*I am not sure I understand your question, but I guess... Counting the ballots, you might have a completely different count for one of the electoral sections than would be expected, for some reason.*" P3 started to think aloud during the interview but confessed he could not come up with an answer: "*[...] Nice [question]! I don't know [how to answer].*" When asked about how the DL module would interact with other models in the system, what sort of connections he could see, P4 answered: "*This is an interesting question. It depends on the configuration of the project, on how the project goes. In theory, the leader of the project is the person who should think about the interaction among the different modules of the system. And then... I don't know. This is a difficult question, to say the truth.*" P5, in turn, was challenged when asked about the most important issues he saw when using a DL model in the context of the proposed BackSys application. After asking the interviewer to repeat the question, he said: "*OK [Pause]. I think that capturing the data [written in the ballots] is the most fundamental problem. I cannot think of anything else.*"

The difficulty and complexity of topics in Part 3 (code *Difficult, Complex, Unknown*) appeared in answers like P2's comment on recognizing *digits* from other marks made by voters in the ballots: "*[We can use pre-processors to filter non-digits] but even in this case, you cannot guarantee that what we get is in fact a digit.*" Further evidence of such complexity can be seen in many of the answers illustrated in the previous paragraph (for example P3's and P4's). An additional illustration of *Difficult, Complex, Unknown* is P1's comment that: "*You have a slot, and people write outside the slot. This makes it more difficult [...] You have to segment the [image] to capture the digit. This is generally a complication of the recognition process.*"

Contents referring to *Development Team and Process* and *Tests, Guarantee, Limitations* were also induced by interview questions in Part 3. To illustrate the participants' answers we can mention P4's comment on team members' responsibilities: "*If I was the project manager, then I'd probably ask someone from Marketing, or Sales, someone from another division, who can understand a bit of statistics and fill the gap between how far I can go to explain this […] and the other end.*" P5 expressed his view on *Tests, Guarantee, Limitations* saying that: "*I guess that it would be possible to issue some output like "this is the number [in the ballot] with such and such level of confidence", right? And then someone would be responsible for accepting the ballot or not, given the uncertainty.*" These examples explicitly denote the participants' tendency to rely on others (referred to as 'someone') to help them deal with what their work with the DL model of BackSys might mean and do in real contexts of use. Another instance of this is P2's comment on how he could help users *Trust* BackSys: "*I could select a set of digit scans from my data base, images that are similar to the one [in question] and have someone decide whether they are similar or not.*"

Additional content about *Trust in Metrics* includes what P1 said he would tell the users, or whoever asked if was BackSys reliable: "*Look, I've gathered all this data, all these variables, and with them I have reached the best possible accuracy.*" P5 provided more explicit metrics: "*With an electronic voting system, nothing less than 100%. [Laughter] Yes, I don't... It's difficult to think of a model that will be correct 100% of the time, but... – or at least 99,9999%, whatever it takes to, given the population of voters, you have less than one wrong vote, or get just one vote wrong.*" P2 also mentioned confidence values: "*So, in view of these kinds of [potential] errors, maybe we could work with the [development] team and look for a way to deliver, along with the computation of votes, some confidence value.*" And P3 commented on the theoretical vs. practical use of statistics: "*The problem is that if you ... if you resort to statistical learning, the theoretical guarantee that we have – unless we have massive amounts of data – the bounds are not very strong. So, what happens in the end is that, as a practice... […] in the general practice of people who develop systems, it's more a matter of empirical than theoretical [decisions].*"

Finally, content about *Models (Learning and Knowledge)* and *Knowledge Limits* include P1's views on DL in practice. He said: "*The implementation of [DL] technology, in practice, has its problems.*" P3 explicitly says that: "*The problem is that many people believe that the model will, by itself, solve the problems.*" And P4 addresses the problem of explainability: "*We know how to explain the model. But it is difficult for us to explain why the model does what it does. In many cases this is not a trivial task...*"

As with Part 1, we studied the relations among the coded meanings illustrated above and produced a conceptual map (Figure 6) with the full articulation (edges) of codes (nodes). Most of the relations articulated are illustrated by the evidence previously provided. We would, however, like to add P5's statement to reinforce our conclusion that users and other affected parties should have trust in metrics: "*I guess somebody must be able to explain this. [laughter] Yeah, I don't know... In the newspapers people say that poll results [have a margin of error] of so many points, above and below. I don't know if the crowds understand this margin of error. Don't know. If they accept what the newspapers say, then maybe they will accept this thing here as well.*" Another noteworthy comment is P3's comments on difficulties, complexity and unknown answers that emerge when you have to think of users and affected parties: "*How annoying to explain this to a lay person, isn't it? [Laughter]*"

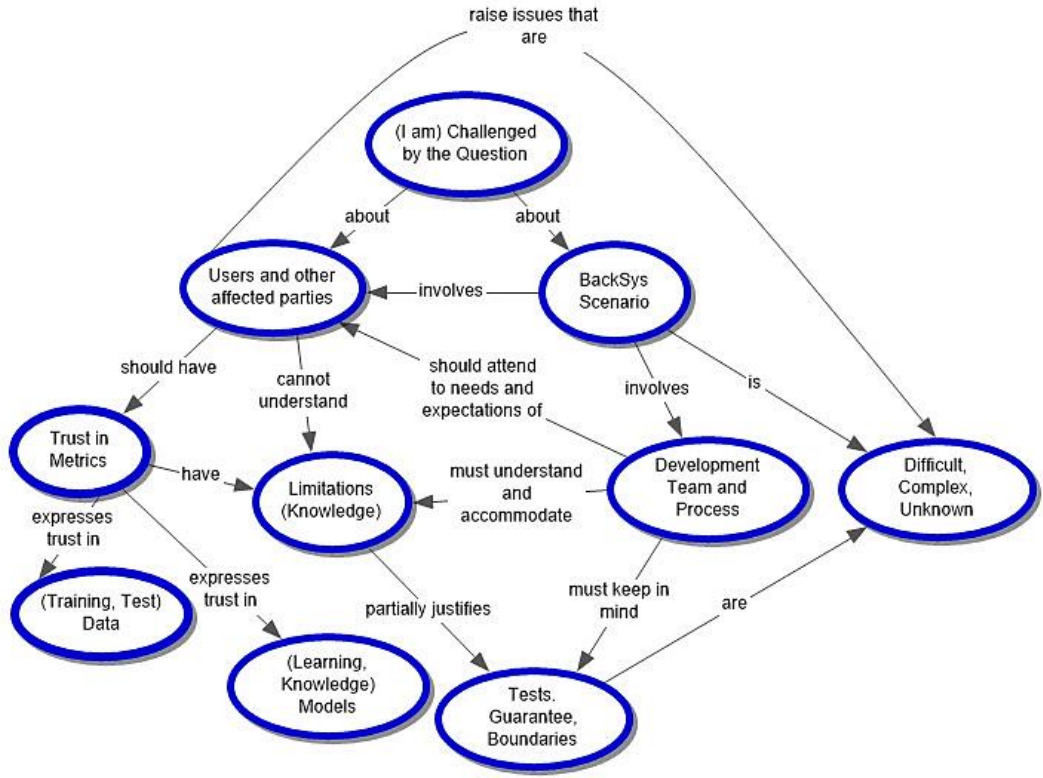

Figure 6: Higher order articulation of themes in Part 3

## 4.4 Conclusions of the Study

The conclusion of the case study emerges when we collate the results of Part 1 and Part 3. Figure 7 shows a curious picture when we compare the participants *spontaneous* (Part 1) reactions to the interview with their reaction to *induced* reflections about the social meaning of DL in real-world applications (Part 3). The picture we get has already been anticipated when we talked about inverted predominant frequency pairs shown in Figure 4. But we now see that the intersecting themes in Parts 1 and 3 were *Trust in Metrics* and *Data (Training and Test)*. The articulations of such themes in both parts of the interview are particularly meaningful. Note that in Part 1, we gathered strong evidence that participants talked about metrics while reflecting about their practice or values. Therefore, the edge that connects *Self-Reflection* and *Trust in Metrics* in Part 1 is labeled as *I have*. In Part 3, as evidence mentioned above has shown, participants manifested their expectation, or hope, that users and other affected parties would trust (have confidence in) metrics that they could show (e.g. accuracy, confidence factors, etc.). We have concluded that *Trust in Metrics* is thus the *link* between participants and the users and other BackSys stakeholders. This connection indicates one of the roots of the *social meaning* that participants see in what they do. The other root is the other intersecting theme, *Data (Training and Test)*, to which the participants pay much attention and, in their view, justifies the stakeholders' trust in metrics. Interestingly, participants talked much more about datasets than about models in Part 1. The conclusion we draw from the intersection of *Data* content in Parts 1 and 3 is that datasets *represent* the users' and stakeholders' context, or at least aspects thereof. The MNIST dataset *represents* digits, which are used to express the voters' intent in the ballots. We don't know, however, as P1 pointed out, whether this representational stance allows us to take MNIST images to stand for ballot *votes*.

The entire map in Figure 7 shows that, during the interview, the participants of our case study have clearly engaged in thinking about the social meaning of DL. Yet, the picture also shows that this sort of consideration is not what comes spontaneously to their minds, *even in the presence of a real-world scenario*. In other words, at *discourse* level (because participants were not *actually* engaged in the development of BackSys) their preference is naturally oriented to technical meanings, rather than social meanings. It would take them more than a realistic scenario description to direct their attention to the social meanings of DL. This is what two participants have clearly told us when asked about their reaction to the material they received. P2 said: "*I remember the proposed scenario, I mean, but... I don't know. As I said I looked at it and I mapped it directly on the tool I already had, which would solve the problem.*" And P3 said: "*I read the scenario, but I picked the keywords: OK, MNIST and Keras! This is what stayed in my mind.*"

Moreover, given the evidence in Part 3, we see that participants said they would rely on team members, project managers, or *someone* to help them deal with social meaning considerations that necessarily arise when developing DL-based technology for applications like BackSys. We conclude that this group would preferably defer considerations about the social meaning of DL, while they could talk at length about many technical aspects of DL. We conclude that the latter is their *comfort zone*, while the former clearly constitutes a *challenge zone*, sometimes explicitly associated with *discomfort* (e.g. P3's "*How annoying to explain this to a lay person, isn't it? [Laughter]*").

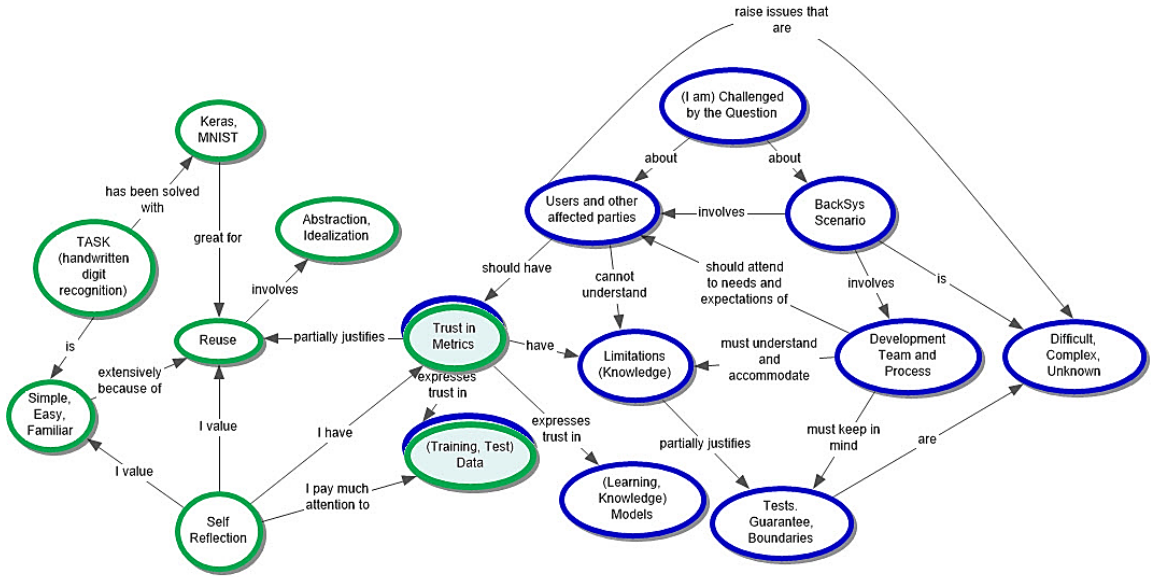

Figure 7: Full higher order articulation of themes in case study interview

One of the powerful findings of our study shows the advantages of qualitative research when investigators want to understand and approach a new object of inquiry. Although the participants were very qualified researchers, four of them holding a PhD degree and one of them in the course of a PhD program, *none* of them viewed the social meaning of DL as a *research topic for themselves*. And, except for P1, who mentioned research about biases in data and algorithms, participants did not acknowledge the social meanings of DL as a research topic at all. All of them referred to it as a development issue that must be resolved when DL is used in practical applications.

This finding led us to look for evidence, in the interviews or elsewhere, of connected facts, which may eventually be investigated as some of the reasons why we found this lack of scientific interest or awareness. In the interviews, the participants' hope that *someone* would have the ability, the skills, the knowledge, to deal with difficult issues related to the social meaning of DL shows that they rely on *mediation* and *mediators* to face the challenges imposed by the case study scenario. In other words, they do not see themselves as moving out of their technical domain, into the social domain. *Someone* will do it. It is also noteworthy that when talking about scientific publications, technical documentation, and other knowledge sources that they value as researchers, in general our participants kept within the boundaries of their discipline (Machine Learning, Deep Learning, Artificial Intelligence). For example, P1 mentioned that: "*These days we have everything in the Internet, you see. So, if there is a new data base, somebody will publish a paper, and usually the author will publish the code [on the Internet], too. And we have thousands of (program) libraries online ... many are for Deep Learning. So, if I want to test a new technique that has been published in a paper, and the code for it is also available, I would take the code and just try to adapt it to, say, another data base that I am using... But I'd stick to the code, no matter which library the author is using*." P1's trust in authors and what they publish can also be traced in P3's comments about handwritten digit recognition tasks. He said: "*For this problem, when *you are [dealing only with] digits, MNIST is certainly a good dataset. Especially for those who have seen Yan LeCun's work. He is the one who created this dataset and what he did with it in the 90's... He did magic, with very little computational power, using this data!"*

These cues into possible explanations for directing scientific attention to *disciplinary* topics, and not directing it to *cross-disciplinary* topics, can be further illuminated by the following excerpts from François Chollet's book on Deep Learning with Keras (2017):

> *"One effective way to acquire real-world experience is to try your hand at machine-learning competitions on Kaggle (https://kaggle.com). The only real way to learn is through practice and actual coding—that's the philosophy of this book, and Kaggle competitions are the natural continuation of this. On Kaggle, you'll find an array of constantly renewed data-science competitions, many of which involve deep learning, prepared by companies interested in obtaining novel solutions to some of their most challenging machine-learning problems. Fairly large monetary prizes are offered to top entrants." (p.337)*

> *"As Kaggle has been demonstrating since 2010, public competitions are an excellent way to motivate researchers and engineers to push the envelope. Having common benchmarks that researchers compete to beat has greatly helped the recent rise of deep learning." (p. 21)*

These two passages (and others, in the book) suggest that benchmarks and competitions stand out among the main incentives and values shared by the DL research community. Note that in Chollet's words, Kaggle competitions are an "effective way to acquire real-world experience." However, we can find examples of socially-justified contexts (see Figure 8), whose evaluation is centered on a receiver operating characteristic curve, plotting predicted probability and observed targets defined by data.

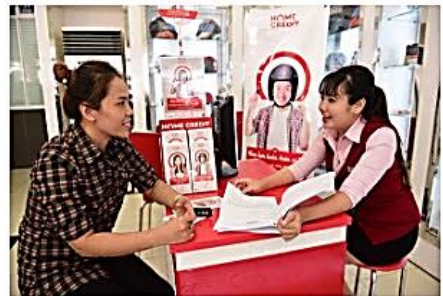

Figure 8: Kaggle's Home Credit Default Risk Competition, In August 2018
(https://www.kaggle.com/c/home-credit-default-risk)

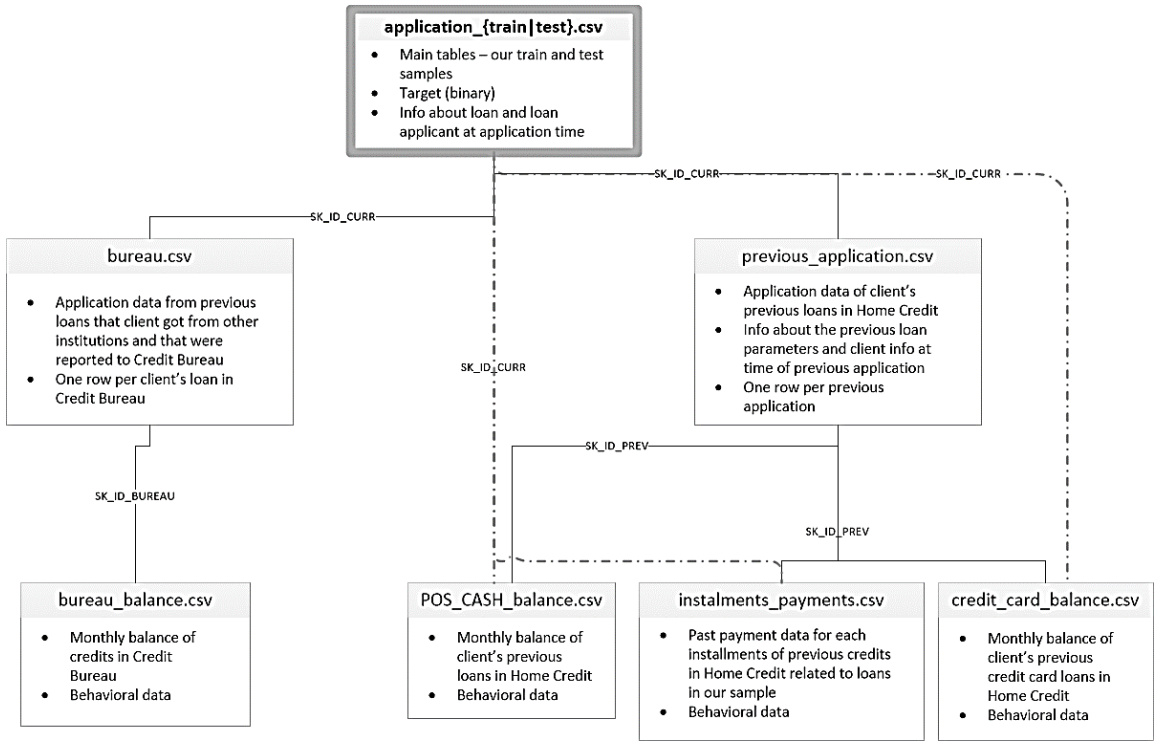

Figure 9: Dataset Structure for the Home Credit Group's Competition on Kaggle,
as of August 2018

The competition's data[1], approved by a global FinTech company (*Home Credit Group*), is structured as shown in Figure 9. It is difficult to see the *social meanings* invoked in the competition's context description (Figure 8) reflected in the data. We can then conjecture that the socio-technical gap we found in our case study is a broader phenomenon than our local case allows us to assert. Therefore, it seems reasonable to conclude that competitions that "motivate researchers and engineers to push the envelope" also motivate them to work with abstract *problem types*, rather than socially contextualized *problem instances*. This is exactly the distinction that we captured with our contrasting perspective codes: task (handwritten digit recognition); problem (ballot image processing); and social scenario (BackSys design, development and use). Given the incentives and values expressed in Kaggle competitions, it is not surprising that the spontaneous reaction of our participants tended towards a *task perspective* (most frequent) and *problem perspective* (second most frequent), rather a *socially contextualized perspective* (least frequent), in Part 1.

### 4.4 XAI Mediation Challenges

The final answer to our research question – How do Deep Learning researchers whose R&D lab work involves the development of real-world applications *relate* to the social meaning of their expert work? – is: they need mediation, in fact, different kinds of mediation. Firstly, they needed the interviewer's mediation to engage in a reflection about the social meaning of their work. Without mediation, their spontaneous reaction was, predominantly, not to take them into consideration. Secondly, when explicitly asked to talk about this topic in Part 3 of the interview, participants frequently relied on *someone* to help them communicate with users and stakeholders of DL applications such as BackSys. Thirdly, their *Trust in Metrics* worked as a *link* (or mediation) between their knowledge and values and the users', as well as *Data (Training and Test)* as a *link* between DL and the real-world context proposed for BackSys.

---

[1] https://www.kaggle.com/c/home-credit-default-risk/data

If we raise the level of interpretation, paying attention to Chollet's views on Kaggle and to some of the active Kaggle competitions at the time we wrote this paper, our results actually show that another kind of mediation may be required. We may have to mediate the cross-fertilization of technical and social research communities. This is because *without* scientific incentives, AI researchers and HCI researchers – to name but the most obvious communities that can extensively collaborate with each other to bridge the socio-technical gap we found with our study – are not likely to come together spontaneously.

The case study and the conclusions reported above allow us to formulate three *XAI Mediation Challenges* to stimulate, at first, collaborative AI-HCI research to bridge the gap revealed by the study. We describe them concisely as three challenging questions, where we choose *one* of many different meanings that the word *mediation* can have. This is the meaning favored by semioticians (Mertz, 2013), who typically discuss *signs* (which in this context we may take as the equivalent of representations) as *mediators* between thought (which in this context we may take as the equivalent of interpretation) and reality (which in this context is socially constructed).

1. **Do training datasets represent AI application users' reality?**
2. **Do training datasets travel across use contexts?**
3. **Does a model learning transfer from one intended use to another?**

These challenges clearly evoke some of the topics mentioned by the participants of our case study, such as data augmentation for transfer learning, for example. They also put the spotlight on *reuse*, a widely disseminated practice in both, research and development contexts. In fact, Chollet (2017) lists three properties the Deep Learning will probably pass on to future AI solutions: simplicity; scalability; and versatility/reusability. About the latter, he says:

> *"Unlike many prior machine-learning approaches, deep-learning models can be trained on additional data without restarting from scratch, making them viable for continuous online learning—an important property for very large production models. Furthermore, trained deep-learning models are repurposable and thus reusable: for instance, it's possible to take a deep-learning model trained for image classification and drop it into a videoprocessing pipeline. This allows us to reinvest previous work into increasingly complex and powerful models. This also makes deep learning applicable to fairly small datasets." (pp. 23-24)"*

The underlying assumption seems to be that DL datasets and models are agnostic to the social meanings that they contribute to create when DL-based real-world applications are used in real-world contexts. When asked about the link between DL and real-world socially meaningful contexts, our participants said that the application system's development team and process should be able to create it. We conclude that in their view the ultimate mediator among all stakeholders involved in the application system's design, development and use is the system itself, that is, *software* where all human meanings are encoded and meet. Hence the importance of challenging DL software components with respect to their representational (or semiotic) stance: what do they stand for (or signify)? To whom? Under what circumstances? And in which respects? Answering these questions, we believe, is a worthy long-term project for researchers interested in XAI. For this reason, we include the XAI mediation challenges listed above as part of our study's results, which demonstrates how qualitative research can contribute to building theories. This is one of the topics we address in the next section.

## 5. Scientific Contribution to XAI and Limitations of this Work

Results and conclusions of qualitative interpretive research yield different contributions to advancing knowledge than is usually the case with quantitative statistical research. As mentioned in preceding sections, qualitative research is suitable for theory-building research projects, and ours is one. We place our lens on a specific segment of theoretical and empirical work to articulate a preliminary socio-technical theory for explainable AI. It must be an incipient theory, given that new developments in AI – especially the work with convolutional neural networks and deep learning – has posed explanation challenges that have only recently begun to be addressed. We still don't have sufficiently robust theories, or at least theories that have been sufficiently tested, to support the generation of explainable DL-based, real-world applications. Therefore, researchers are not only free to choose how to start building them, but actually stimulated to do so, in face of rapidly spreading *black box* AI being used in socially sensitive applications such as credit default risk analysis, for example (see the Kaggle competition illustration in Figure 8).

Our non-exclusive choice, among many possibilities, is to theorize about socio-technical gaps in AI, starting from pragmatic theories of discourse and communication that supported the automatic generation of explanations for expert systems in the 1980's and 1990's (Cawsey, 1992; Hovy, 1987, 1990; Maybury, 1992; Moore, 1994; Paris, 1991; Scott & de Souza, 1990). This body of work has shown that appropriate explanations are generated under pragmatic constraints, which include among other factors an adequate understanding of an intelligent system's users' goals, beliefs, context, etc. We thus *anticipated* that, regardless of the kind or degree of *explainability* that we will be able to achieve with DL, to say *anything* at all about how DL-based AI systems work or why they behave the way they do, we must ourselves – as contemporary AI researchers – be able to *relate* to contemporary intelligent systems users' goals, beliefs, contexts, etc. Some may dispute that this is a *theoretical* formulation, rather than a *common sense* one. Indeed, it is a very basic theoretical requirement that explanation-givers be able to relate to explanation-receivers in order to take the receivers' goals, beliefs and context into consideration. But this is precisely where the contribution of our work resides. When we investigated this requirement empirically, with a small group of DL researchers whose job in an R&D lab involves the development of real-world applications, we found that it is a considerable challenge for them to *relate* to real-world application users, even in a context where *they themselves* would be users of the proposed system. Remember that all participants of our study were thoroughly acquainted with the socio-cultural value of computer-aided voting systems because, as voting citizens of the country of reference for our case study's scenario, they are users of the existing EVS and, therefore, also legitimate users of BackSys.

The fact that our participants cannot directly relate to users, that they need different kinds of mediation to do it, is an important empirical counterfactual in our theoretical construction. It means that we must theorize about mediation in our way to theorizing about explanations for contemporary AI applications. Since those who build DL kernels of AI applications cannot easily relate to users and explicitly call for someone else's intervention, we may be heading towards the same old software engineering problems that produced a large volume of low-usability systems. We know that high-usability systems depend on integrating usability engineering and software engineering into the same process (Seffah et al., 2005; Seffah & Metzker, 2004). In other words, we now know that usability is not a later addendum to a user-agnostic software development process. Quite contrarily, usability can only be reached if users are taken into consideration in all steps of the software development process. Therefore, because *explainability* is clearly a *usability* requirement for contemporary AI systems, mediation and its challenges constitute a key research topic for empirical and theoretical inquiry to bridge socio-technical gaps in AI. Thus, the scientific contribution of our work is to diagnose the problem, in a richly contextualized occurrence.

Nevertheless, our study has essential limitations, which must be clearly stated. The main one is that its results and conclusions cannot be generalized to the larger research community who is interested in or involved with the development of real-world DL applications. Although we conjecture that, given the previously mentioned evidences, such as Chollet's book (2017), and in current competitions appearing in Kaggle's website, our findings may reach considerably farther than the limits of our case study impose. We cannot know it now. Hence, further research is necessary to confirm this.

A second limitation, seldom emphasized in qualitative research reports, is that the replicability of this study is impossible, and not only because other participants, working in another context, under different conditions, coming from a different micro or macro culture will most probably respond differently to the proposed *hand-on* activity and interview. Our case study participants *themselves* have changed views during the interview. So, if we interviewed them once again we would necessarily have different answers. The interview was in itself an intervention. They were, therefore, transformed by it. One of our participant's statement during the member-checking validation session clearly illustrates the non-replicability of our research. He said: "*I just wanted to add that I was not aware at all about my biased vision in what concerns the research that I conduct. It is very clear now that all my answers have been driven by technical aspects of the problem and not as much by social impact that such research could have.*"

A third important limitation of this work is that we detected a *socio-technical gap* without having heard what the users of DL applications and other affected parties have to say about it. We only heard the technical side, not the social side. Indeed, this limitation clearly calls for a similar study with ML systems users. Yet, the strength of our conclusions lies in the fact that we did not start the research looking for a socio-technical gap. The gap was explicitly exposed by the technical participants themselves, which is a way of legitimizing the presence of the gap. If one group feels that they cannot reach the other, and that they need mediation to do so, the gap is real even if the other group has the necessary resources to bridge it.

Finally, an additional limitation in the research presented here is that we have concentrated our XAI Mediation Challenges on the representational aspects of mediation. Another equally important aspect of mediation, i.e. the role of mediators in communication processes, helping messages get across from original senders to final recipients, has not been elaborated in this paper. We are still investigating the challenges that we can contribute in this particular respect, with a special emphasis on semiotic mediation (Mertz, 2013), which can bring together their communicative and representational dimensions.

Therefore, the above limitations mean that our main contribution is, indeed, to build new theoretical constructs – the concept of mediation, along with its various roles and facets in terms of narrowing the socio-technical gap in the development of real-world DL applications – which interested XAI researchers can use to design and execute further studies, following quantitative or qualitative, predictive or non-predictive orientations.

## 6.  Concluding Remarks

In this paper, we presented a case study carried out with a group of Deep Learning experts. We included a detailed description of our research design, showing how the case was defined and what methods and techniques were used to analyze the evidence collected at different stages of our project.

Our research question was: **how do Deep Learning researchers whose R&D lab work involves the development of real-world applications relate to the social meaning of their expert work?** The social meaning of AI (what it means *to others*) is a topic that our research community must investigate, given the societal pressure for more transparency and data owners' control regarding how personal information is used in data-driven machine learning applications.

Our qualitative research centered on interviews with five expert researchers from a R&D laboratory focused on the delivery of industrial-strength AI technologies. The findings of the study revealed a socio-technical gap between the participants´ perspective on their work with DL, compared to their perspective on how their work might affect actual users and other affected parties in a real-world DL application. An in-depth thematic analysis of this group´s collective discourse also showed that their research values and incentives are apparently at odds with the kind of multi-disciplinary research initiatives that would be required to bridge this gap. For example, benchmarks and competitions that are used to validate and advance scientific progress in machine learning, typically abstract away the contextual details that turn scientific and technical knowledge into socially meaningful solutions to real-world problems. Additionally, we detected that mediation was a central theme in the case study, from mediation needed to help experts communicate with users, to that needed to help AI researchers recognize the scientific challenges that social values and meanings may bring to their own research practice.

We proposed three XAI Mediation Challenges as a contribution to the broader collective effort, which research communities from many different areas must make, to advance the state of the art in the field of explainable AI. They concentrate on the representational aspects of mediation, starting from how datasets *represent* the world and context of those who use ML applications. Datasets represent a link, or mediation, between technology and social reality. Therefore, some of the techniques being investigated in ML, such as data augmentation and transfer learning, may have to be revisited with a socio-technical perspective in mind.

As is typical with qualitative research, our study raises numerous follow-up questions, many of them commented in this paper. Part of our future work will focus on the relation between training and test datasets used to develop the AI component of real-world applications and these applications´ users´ context. Another part, as mentioned among the current limitations of our research, is to explore communicative mediation challenges, from a semiotic perspective.

**Acknowledgements**

The authors thank the anonymous participants of this study for their generous consent and invaluable contribution to the authors' research. De Souza, additionally thanks IBM Research for supporting her sabbatical leave, and the Brazilian National Council for Scientific and Technological Development (CNPq) for partially supporting this research (Grant # 304224/2017-0).

## Appendix A: Survey Questions

For how long have you been working with Machine Learning?

- [Less than 1 year/1 to 3 years/4 to 6 years/7 years or more]

What was/has been your motivation or reason to work with ML?

- [open question]

Can you briefly tell us about how you have learned (or have been learning) ML?

- [open question]

Do you work with neural networks and/or deep learning? (Feel free to comment on your answer.)

- [Yes, with both./Not exactly./Not at all.]
- [Request for extra comments]

What ML library/framework do you usually work with in your projects? (Please check all the applicable.)

- [Keras/Caffe/TensorFlow/Other (please specify)]

Can you tell us of TWO POSITIVE ASPECTS (ADVANTAGES) of the library/framework that you work with?

- [open question]

Can you tell us of TWO NEGATIVE ASPECTS (DISADVANTAGES) of the library/framework that you work with?

- [open question]

In view of advantages and disadvantages listed above, how do you explain your choice of technology?

[open question]

## Appendix B: Scenario and Hands-on Activity

*Scenario:*

The form, or ballot, has blank squares next to each elective post (president, governor, senator, congressman and state deputy). The voter must write only one digit per square, forming the number that corresponds to their chosen candidate (or party, in the case of congressmen and state deputies). For example, in Figure 1B a written vote is shown for the presidential candidate registered as number "18" (the name of the candidate, such as in the electronic ballot platform, is not informed). BackSys automatically recognizes the written votes, counts them and sends voting totals to another system responsible for producing election final results. The BackSys technical details have already been presented to the design and development team in a prior meeting. One of its main characteristics is the input interface that includes a scanner to capture ballots images. Once the ballot image is captured, a citizen's vote is classified as valid or invalid. According to current legislation, valid votes (for each elective post) are those that: are recognized and correspond to a candidate (or party) number; or for which all the ballot's squares are blank. Invalid votes are: those with recognizable numbers that do not correspond to any candidate; or votes with any other written content not recognized as a candidate's number or party. Only valid votes count for candidates or parties.

*Hands-on activity:*

Your task is to create a deep learning model to recognize hand-written digits, which is a central component for BackSys. To this end, you will use Keras, with which you are familiar. The dataset MNIST is already at your disposal. If you think you need further information or resources, please talk to the researcher that contacted you about this study. When you are satisfied with your model, or whenever you want or have to stop the work for any other reason, save your model (as a *.py file) and send it to the researcher who has contacted you to complete the hands-on activity. The researcher will want to hear your comments and opinions on some specific aspects of what you have just done. So, we ask you, please, to give us a short interview, which we will schedule when you submit your model.

Figure B1: Sketch of the ballot that BackSys needs to process.
       A → empty ballot
       B → ballot with manually registered vote

## Appendix C: Interview Guide

### About the Scenario and Hands-On Activity

Collect impressions, doubts, ideas, thoughts which occurred to the participants when they read the hands-on material (scenario and activity description).

*Guiding questions:*

- What did you do when you got the hands-on material? Where did you start? How did you approach the proposed task?

- Which aspect(s) in the scenario caught your attention? Why?

### About Building Models Using Keras

Talking with participant about the model built with Keras, what the participant considers important or relevant in the code, how was the coding process

*Guiding questions and discussion items:*

- Can you highlight what is important for you in your code?

- How did you get to this model code? Where did you start? Were there intermediary steps? Are you satisfied with your code?

- If the participants haven't mentioned the items below, ask them to comment on:
    - Parameters (batch size, epochs)
    - Net architecture (how many layers, which types of layers, configuration details of each layer)
    - Model training (loss function, optimizer, metrics)

### Relating the Model with the Scenario

Going back to the scenario, as the participants to comment on:

- What they think their work/role would be like in the proposed scenario, and where they would start

- Considering the complete scenario in the hands-on activity, how would the participants proceed? For example, would Keras and MNIST be the participant's choice for that scenario of digit recognition on voting ballots? Why?

- What kind of interaction(s) would the participants (who is responsible for building the deep learning model) expect to have with the rest of the BackSys development team? What kind of input would the participants expect to get from other team member, what kind of output do they think they would have to produce? What would their collaboration be like?

Considering the use of BackSys in the participants' country's election:

- What do you think BackSys would be prepared to do or not?

- What would the system do better (positive points) than manual voting registration and why? What would it do worse (negative points)?

- If BackSys was a real, existing system, and if you were part of the development team, what kind(s) of doubts do you anticipate that users and stakeholders might have regarding the system's behavior? (Think, for example, of voters, political parties, the Supreme Court).

- What kind of explanation about BackSys someone could ask? Why?

# Appendix D: Codebook

*Survey Codebook:*

- **Experience**
  - Novice
  - Not a Novice
  - Expert
- **Tools**
  - Keras
  - Caffe
  - TensorFlow
  - Other Tool
  - None of 3 Named Tools
- **Motivation**
  - Problem Solving, Applications
  - Mathematical Modeling, Science
  - Other Motivation
- **Training**
  - Structured University Education
  - Online Training
  - **Other Ways of Learning**
    - Self Taught / Self Initiative / Documentation
    - In Job Training / Peers
- **Technology**
  - Range and Flexibility
  - Model / Program Libraries
  - Ease of Use
  - Software Documentation
  - Programming Capabilities
  - Engaged Online Community
  - Open Source or Proprietary
  - Constantly Evolving
  - Performance
  - Cost/Benefit
  - Recommendation
  - Good for Research
- **Domain**
  - Image Processing
  - Remote Sensing
  - Pattern Recognition
  - High Performance Computing (HPC)

*Case Study Codebook:*

- **Reaction to the Interveiw**
  - Self-Reflection
  - Revision of Previous Statement
  - Challenged by the Question
- **Perspective**
  - Social Scenario of the Interview
  - Parallel Social Scenario
  - Problem to be Solved (Image Processing)
  - ML Task (Handwritten Digit Recognition)
- **Attitude and Relation**
  - Simple, Easy, Familiar
  - Difficult, Complex, Unknown
  - Interesting, Challenging
  - Unintesting, Boring
  - Can be controlled
  - Cannot be controlled
- **Trust**
  - Trust in people
  - Trust in artifacts
  - Trust in metrics
- **Machine Learning**
  - Keras MNIST
  - Data (Training and Test)
  - Models (Learning and Knowledge)
  - Tools
  - Experimentation, Tuning, Optimization
  - Knowledge Limits
  - Small Data, Transfer Learning, Data Augmentation
  - Empirical proof
  - Biases
  - Publications (Technical and Scientific)
- **Software Development and Stakeholders**
  - Abstraction, Idealization
  - Reuse
  - Testing, Guarantee, Limitations
  - Development Team and Process
  - Users and Other People Affected by the System

MEDIATION CHALLENGES AND SOCIO-TECHNICAL GAPS FOR EXPLAINABLE DEEP LEARNING APPLICATIONS## References

Abdul, A., Vermeulen, J., Wang, D., Lim, B. Y., & Kankanhalli, M. (2018). Trends and Trajectories for Explainable, Accountable and Intelligible Systems: An HCI Research Agenda. In *Proceedings of the 2018 CHI Conference on Human Factors in Computing Systems* (pp. 582:1–582:18). New York, NY, USA: ACM. https://doi.org/10.1145/3173574.3174156

Binns, R. (2017). Algorithmic Accountability and Public Reason. *Philosophy & Technology*. https://doi.org/10.1007/s13347-017-0263-5

Biran, O., & Cotton, C. (2017). Explanation and justification in machine learning: A survey. In D. W. Aha, T. Darrell, M. Pazzani, D. Reid, C. Sammut, & P. Stone (Eds.), *IJCAI-17 Workshop on Explainable AI (XAI)* (pp. 8–13). Melbourne, Australia. Retrieved from http://home.earthlink.net/ dwaha/research/meetings/ijcai17-xai

Blandford, A. (2013). Semi-structured qualitative studies. Interaction Design Foundation.

Blandford, A., Furniss, D., & Makri, S. (2016). Qualitative HCI Research: Going Behind the Scenes. *Synthesis Lectures on Human-Centered Informatics*, *9*(1), 1–115. https://doi.org/10.2200/S00706ED1V01Y201602HCI034

Braun, V., & Clarke, V. (2006). Using thematic analysis in psychology. *Qualitative Research in Psychology*, *3*(2), 77–101.

Brusilovsky, P. (1994). Explanatory visualization in an educational programming environment: Connecting examples with general knowledge. In B. Blumenthal, J. Gornostaev, & C. Unger (Eds.), *Human-Computer Interaction* (pp. 202–212). Berlin, Heidelberg: Springer Berlin Heidelberg.

Cawsey, A. (1992). *Explanation and interaction: the computer generation of explanatory dialogues*. Cambridge, Mass: MIT Press.

Chollet, F. (2017). *Deep learning with python*. Manning Publications Co.

Creswell, J., & Clark, V. (2017). *Designing and conducting mixed methods research*. Sage publications.

Creswell, J., Hanson, W., Clark, V., & Morales, A. (2007). Qualitative research designs: Selection and implementation. *The Counseling Psychologist*, *35*(2), 236–264.

Denzin, N. K., & Lincoln, Y. S. (Eds.). (2005). *The SAGE handbook of qualitative research* (3rd ed). Thousand Oaks: Sage Publications.

Diakopoulos, N. (2016). Accountability in Algorithmic Decision Making. *Commun. ACM*, *59*(2), 56–62. https://doi.org/10.1145/2844110
35